\documentclass[runningheads]{llncs}
\usepackage{graphicx}
\usepackage{amsmath}
\usepackage{amsfonts}
\usepackage{color}

\newcommand{\letter}[1]{`{\tt #1}'}
\newcommand{\letterc}[1]{`{\tt #1},'}
\newcommand{\letterp}[1]{`{\tt #1}.'}

\begin{document}
\title{Toward Defensive Letter Design}%
%
\author{Anonymous submission}
\author{Rentaro Kataoka\inst{1}\and
Akisato Kimura\inst{2}\and
Seiichi Uchida\inst{1}\orcidID{0000-0001-8592-7566}}

\authorrunning{R. Kataoka et al.}

\institute{Kyushu University, Fukuoka, Japan \email{rentaro.kataoka@human.ait.kyushu-u.ac.jp}
\and
NTT Corporation, Kanagawa, Japan\\
}
\maketitle              
\begin{abstract}
A major approach for defending against adversarial attacks aims at controlling only image classifiers to be more resilient, and it does not care about visual objects, such as pandas and cars, in images. This means that visual objects themselves cannot take any defensive actions, and they are still vulnerable to adversarial attacks. In contrast, letters are artificial symbols, and we can freely control their appearance unless losing their readability. In other words, we can make the letters more defensive to the attacks. This paper poses three research questions related to the adversarial vulnerability of letter images: (1) How defensive are the letters against adversarial attacks? (2) Can we estimate how defensive a given letter image is before attacks? (3) Can we control the letter images to be more defensive against adversarial attacks? For answering the first and second questions, we measure the \emph{defensibility} of letters by employing Iterative Fast Gradient Sign Method (I-FGSM) and then build a deep regression model for estimating the defensibility of each letter image. We also propose a two-step method based on a generative adversarial network (GAN) for generating character images with higher defensibility, which solves the third research question.
\end{abstract}
%
%
\section{Introduction\label{sec:intro}}
{\em Adversarial attack}, one of the hot topics in recent machine learning research, is a technique to give artificial distortions or deformations to a sample so that a classifier misrecognizes the sample. The main focus of adversarial attack research is to analyze the vulnerability of deep neural network (DNN)-based classifiers. Many attack algorithms have been developed so far~\cite{Chakraborty2018} and applied to various types of data, such as images, speech signals, and texts. Especially for images, we can find various attack algorithms~\cite{Akhtar2018,Xu2020,Kaviani2022,Long2022,Machado2021}. In a recent survey~\cite{Machado2021}, major image attack algorithms are classified into gradient-based, transfer/score-based, decision-based, and approximation-based algorithms.\par
%
{\em Adversarial defense}, also a hot topic, is a technique to make the DNN-classifiers resilient to adversarial attacks by utilizing the results of analyzing adversarial vulnerability.  Like the attack algorithms, many defense algorithms have been proposed, and they are classified into gradient masking (including adversarial training and defensive distillation), auxiliary detection models, statistical methods, preprocessing techniques, classifier ensembles, and proximity measurements~\cite{Machado2021}.
\par
%
\begin{figure}[t]
\centering
\includegraphics[width=\textwidth]{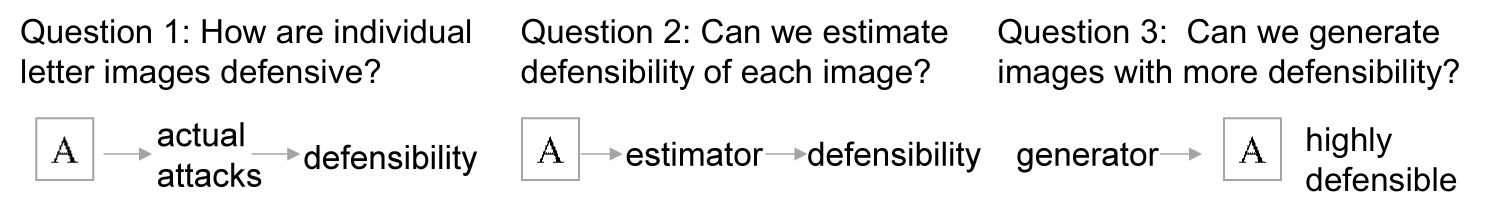}\\[-2mm]
\caption{Three research questions on adversarial attack and defense of letter images.} \label{fig:RQs}
\end{figure}
%
Meanwhile, this paper focuses on the {\em defensibility} of {\em letter images} (such as \letter{A}-\letter{Z}); in other words, we focus not on the defensibility (or vulnerability) of classifiers but on the defensibility of classification subjects, i.e., letters. Specifically, we set three research questions on adversarial attacks and defenses of letter images, as summarized in Fig.~\ref{fig:RQs}. These questions are specific to letter images because they have unique characteristics as {\em artificial symbols designed by humans for humans}, whereas general visual objects, such as ``pandas'' and ``cars,'' do not. In the following, the three questions are detailed with their purposes.
\par
%
\begin{itemize}
\item {\bf Question 1: How defensive are the letters against adversarial attacks?}\  
We, humans, can read letters that are attacked by various distortions and deformations, such as perspective distortions, handwriting fluctuations, partial occlusions, blurs, dot noise, and styles (i.e., typeface designs, such as Romans, Grotesque, and Geometric). Therefore, for ``human classifiers,'' letter images are already robust to these attacks. In other words, the letter images are designed to keep their discriminability (at least for humans) under various attacks. In contrast, we do not know how letter images might have high defensibility against attacks on DNN-based classifiers.
\item{\bf Question 2: Can we estimate how defensive a given letter image is before attacks?}\  
Different letter images (i.e., letter appearances) may have different strengths against the attacks. If so, the strength should depend on the appearance of the letter images. In other words, we need to confirm the possibility of estimating the defensibility of each image without actual attacks.
%
\item{\bf Question 3: Can we control the letter images to be more defensive against adversarial attacks?}\  
As noted above, standard approaches for defending against adversarial attacks try to control only image classifiers to be more resilient, and they do not care about visual objects (such as ``pandas'' and ``cars'') in images. In other words, visual objects are just left to be attacked without any defensive actions. However, letters are artificial symbols, and we can freely control their appearance unless losing their readability. In other words, we can {\em design} the letter images to be more defensive against the attacks. Over thousands of years of our history, humans have controlled the letters from cuneiform to the modern Latin alphabet so that they would be defensible (i.e., readable) to human classifiers even under the above distortions and deformations. Now, it seems worthwhile to consider designing the letters to be more defensive even against adversarial attacks against non-human classifiers, i.e., DNN-based classifiers.\par
\end{itemize}

Our main contributions are to give possible answers to the above three questions, which are summarized as follows.
\begin{itemize}
\item {\bf Measuring defensibility by attacking}:\  
The first contribution is to measure the actual defensibility of each letter image $\mathbf{x}$ by repetitive attacks. We define the defensibility of a letter image $\mathbf{x}$ as $k(\mathbf{x})=k$ when $\mathbf{x}$ is first misrecognized only after $k$ attacks. As an attacking algorithm, we use Iterative Fast Gradient Sign Method (I-FGSM)~\cite{Kurakin2017}, one of the most popular repetitive attacking algorithms. We also observe the differences in the defensibility between letter classes.  
\item {\bf Estimating defensibility by regression}:\  
The second contribution is to realize a deep regression model that can estimate the defensibility $k(\mathbf{x})$ of each letter image without actual attacks. More formally, we build a nonlinear function $\hat{k}=\hat{k}(\mathbf{x})$ that obtains an estimate $\hat{k}$ of the defensibility of a letter image $\mathbf{x}$. If we can obtain such a regression model $\hat{k}(\cdot)$ accurately, it will experimentally prove a mutual relationship between letter shapes and their defensibility.
\item {\bf Generating defensive letter images}:\ 
The third contribution is to propose a two-step method for generating letter images with higher defensibility based on a generative adversarial network (GAN). After pre-training the model with a standard GAN framework for generating readable letter images, we further train the generator with a new loss function to increase the defensibility of generated letter images.
\end{itemize}
\par
To the authors' best knowledge, it is the first attempt to understand the defensibility (or vulnerability)  of letter images from the viewpoint of adversarial attack and defense. Although this defensibility is evaluated by a non-human classifier (i.e., a DNN-based classifier), we expect that the results will help our future work to understand the robustness of letters for humans. 
\par
\section{Adversarial Attacks\label{sec:review}}
As noted in Section~\ref{sec:intro}, many algorithms for adversarial attacks and defenses have already been proposed. Since the main focus of this paper is not to propose some new algorithm for adversarial attacks and defenses, we will not go into their details. For readers interested in them, please refer to recent surveys, such as Mechado et al.~\cite{Machado2021}.\par
%
In our trial, we use Iterative Fast Gradient Sign Method (I-FGSM)~\cite{Kurakin2017}. We employ I-FGSM because of two reasons. First, I-FGSM is a general attack method and gives the basis of state-of-the-art attack methods~\cite{Wu2021}~\cite{Wang2021}. Second, I-FGSM is a repetitive attack method, and thus suitable for quantitatively measuring the defensibility of each sample. Using different attack methods might produce different results than this paper. (In fact, any attempts at adversarial attack and defense cannot escape from the dependency on the attack method.) We expect that choosing I-FGSM makes our results as general as possible.\par
I-FGSM can be seen as an extension of a classical gradient-based attack called Fast Gradient Sign Method (FGSM)~\cite{Goodfellow2015}. 
FGSM uses the gradient $\nabla_\theta J(\theta, \mathbf{x}, y)$ with respect to the target model to be attacked for generating adversarial examples $\mathbf{x}'$, where $\theta$ is the parameters of the target model, $\mathbf{x}$ is the input example, $y$ is the ground-truth class label for $\mathbf{x}$, and $J(\theta, \mathbf{x}, y)$ is the loss function for training the model. Specifically, the adversarial example $\mathbf{x}'$ generated from $\mathbf{x}$ by FGSM is expressed as:\vspace{-2mm}
\begin{equation}
\mathbf{x}'=\mathbf{x}+ \epsilon\cdot\mathrm{sign}\left(\nabla_\theta J(\theta, \mathbf{x}, y)\right),
\label{eq:FGSM}\vspace{-2mm}
\end{equation}
where the second term on the right side is an adversarial perturbation, and $\epsilon$ controls the amplitude of the perturbation. The perturbed example $\mathbf{x}'$ is expected to give a larger loss value than $\mathbf{x}$, and thus $\mathbf{x}'$ can behave as an adversarial example.\par
%
Roughly speaking, I-FGSM generates adversarial examples by repeating FGSM of Eq.~(\ref{eq:FGSM}). By repetitive attack operations, I-FGSM will generate ``more adversarial'' examples. In fact, we can repeat Eq.~(\ref{eq:FGSM}) until $\mathbf{x}'$ is misrecognized as a different class from the original class of $\mathbf{x}$. Formally, this repetitive attacking process is described as follows:\vspace{-2mm}
\begin{equation}
    \mathbf{x}'_{t+1}=\mathbf{x}'_{t} + \epsilon\cdot\mathrm{sign}\left(\nabla_\theta J(\theta, \mathbf{x}'_{t}, y)\right),
    \label{eq:I-FGSM}\vspace{-2mm}
\end{equation}
where $t$ is the number of attacking iterations.
\par

\par
As noted in Section~\ref{sec:intro}, all letters are designed by humans. In the past, there were typefaces for not only human classifiers but also computer classifiers. For example, OCR and MICR fonts were designed for OCR (optical character reader) systems and MICR (magnetic-ink character reader) systems, respectively. Letter shapes of OCR fonts are designed to have clear discriminability. For example, \letter{O} of an OCR font called ``OCR-A'' (developed in 1968) looks like \letter{$\diamond$} to keep discriminability from \letter{0} (zero). Literature~\cite{uchida2007} designs characters where computer-readable class information is embedded in a special way for perspective-invariant information retrieval. These classic attempts are pioneering works in which defensive letter designs were made for non-human classifiers against certain types of attacks.
\par
\section{Measuring Defensibility by Attacking \label{sec:attack}}
\subsection{Attacking Letter Images by I-FGSM}
To answer Question 1 in Section~\ref{sec:intro} ({\it ``How are the letters defensive against adversarial attacks?''}), we attack letter images by I-FGSM and then measure their actual defensibility. Specifically, we first train a CNN for letter classification using training and validation sets of letter images. Then, we classify the letter images in a test set and discard the misrecognized images. Finally, we run I-FGSM for each non-discarded image $\mathbf{x}$. The number of attacking iterations until misrecognition for $\mathbf{x}$ is determined as the defensibility $k(\mathbf{x})$. Note that $k(\mathbf{x})>0$ for any correctly recognized (i.e., non-discarded) image $\mathbf{x}$.\par
%
We set the parameter $\epsilon$ of the step size in FGSM at $0.02$ through a preliminary experiment. Although there is neither a theoretical nor experimental criterion to determine the value of $\epsilon$, it is inappropriate to set it at a large value. This is because most images are misrecognized just by a one-time attack with a large $\epsilon$, and we cannot observe the different defensibility among letter images for this case. Setting $\epsilon$ at a very small value is also inappropriate because it requires too many iterations until misrecognition. Considering these points in preliminary experiments, we set $\epsilon=0.02$ as a reasonable compromise. Under this setting, the defensibility $k$ fell in the range $[1,32]$ for any image used in the experiment.
\par 
\begin{figure}[t]
\centering
\includegraphics[width=0.9\textwidth]{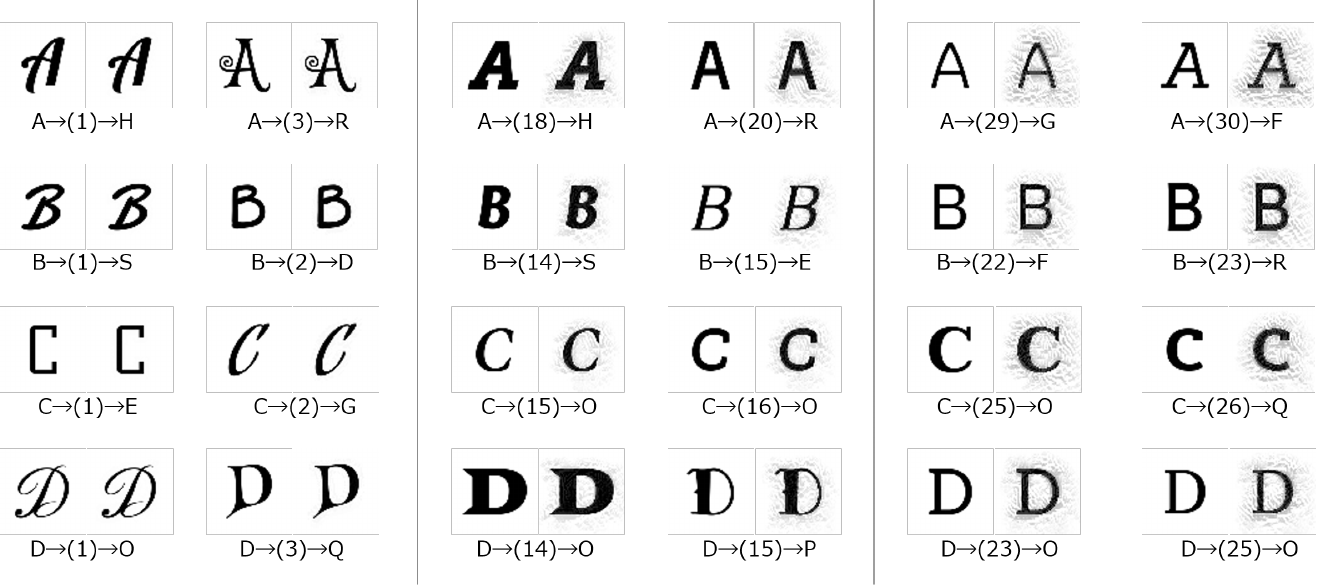}\\[-5mm]
\caption{Examples of attacked letter images. For each example, the original image and the attack result are shown. The notation ``$\mathtt{A}\to$(1)$\to\mathtt{H}$'' means that the original class is \letterc{A}  the number of attacking operations (i.e., defensibility $k$) is 1, and the misrecognized class is \letterp{H}} \label{fig:attacked-examples}
\end{figure}
\begin{figure}[t]
\centering
\includegraphics[width=0.88\textwidth]{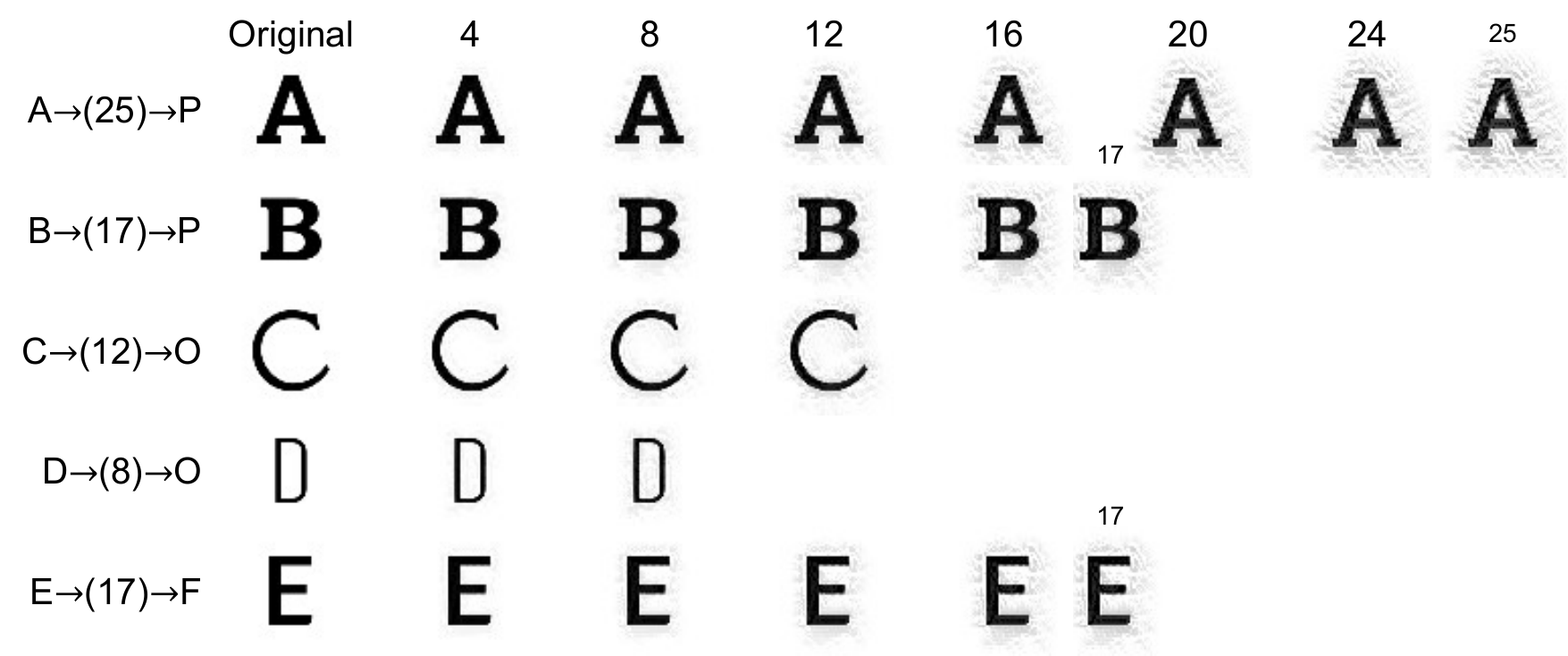}\\[-3mm]
\caption{The attacking process by I-FGSM.} \label{fig:attacked-sequence}\vspace{-5mm}
\end{figure}

\subsection{Attack Experiment\label{sec:attack-experiment}}
\subsubsection{Letter Image Dataset}
We used Google Fonts to collect the letter images to be attacked. Specifically, we used images of 26 alphabets (\letter{A}-\letter{Z}) of 3,124 different fonts from Google Fonts. Each letter image is a $64\times 64$ binary image with $\pm 1$ pixel values. All the images ($26\times 3,124$) are split into three font-disjoint datasets, i.e., training, validation, and test sets, and they contain about 30,000, 4,000, and 30,000 images, respectively. \par 
\subsubsection{CNN to Be Attacked}
A simple CNN $C$ with two convolutional layers and two fully-connected layers is trained for the 26-class classification task. Each convolutional layer is accompanied by ReLU activation and max-pooling. In its training process, the negative log-likelihood loss is used and minimized by AdaDelta. The training process is terminated by the standard early stop criterion, whereby training stops if the validation loss does not decrease over the next 10 epochs. After training, the training, validation, and test accuracies are 98.3\%, 95.1\%, and 95.7\%, respectively.
\subsubsection{Attacked Examples}
The correctly recognized images in the test set (i.e., about $28,700\sim 30,000\times 95.7\%$ images) are attacked by I-FGSM. Fig.~\ref{fig:attacked-examples} shows the examples of attacked letter images. Among the three columns (separated by vertical lines), the left, middle, and right columns show low defensive (i.e., fragile), moderate defensive, and highly defensive (i.e., robust) cases, respectively. Below each image, its defensibility $k$ and the misrecognized class (after $k$ attack operations) are shown.\par
%
Those examples suggest that the letter images by decorative fonts tend to have smaller defensibility; that is, they are often fragile against adversarial attacks. This is because they are often outliers or located far from the center of the distribution of each letter class and, therefore, nearby the distribution of a neighboring class. Consequently, even with a small number of attack operations, they move into the neighboring class. In fact, 
in most cases, the misrecognized class resembles the original class, such as \letter{A} $\leftrightarrow$\{\letterc{H} \letter{R}\} and \letter{B} $\leftrightarrow$\{\letterc{S} \letterc{D} \letter{E}\}. \par
In contrast, the letter images by orthodox fonts have large defensibility $k$ since they are located around the center of the distribution. By many attack operations, those robust letters are often misrecognized as a class with a largely different shape, such as \letter{A} $\leftrightarrow$\{\letterc{G} \letter{F}\} and \letter{B} $\leftrightarrow$ \letterp{F} 
\par
%
Fig.~\ref{fig:attacked-examples} also shows that the letter images with more attacks ($>15$) seem like a blurred version of their original image. In other words, the attacks appear as gray pixels nearby the original strokes and do not like random salt-and-pepper noise over the entire image region. This result proves that the computer classifier misrecognizes the letter images just by those blurring-like attacks;  in other words, more harsh attacks, such as drastic stroke shape deformations, are not mandatory. See Section~\ref{sec:discussion-attack} for a discussion about the type of attack.
\par
%
Fig.~\ref{fig:attacked-sequence} shows the sequence of the attacking operations by I-FGSM. As seen in the case of \letterc{A} the blurred area around the stroke expands gradually and monotonically along with attacking iterations, $k$. It is also confirmed that I-FGSM does cause neither abrupt changes nor drastic stroke shape deformations.

\begin{figure}[t]
\centering
\includegraphics[width=0.95\textwidth]{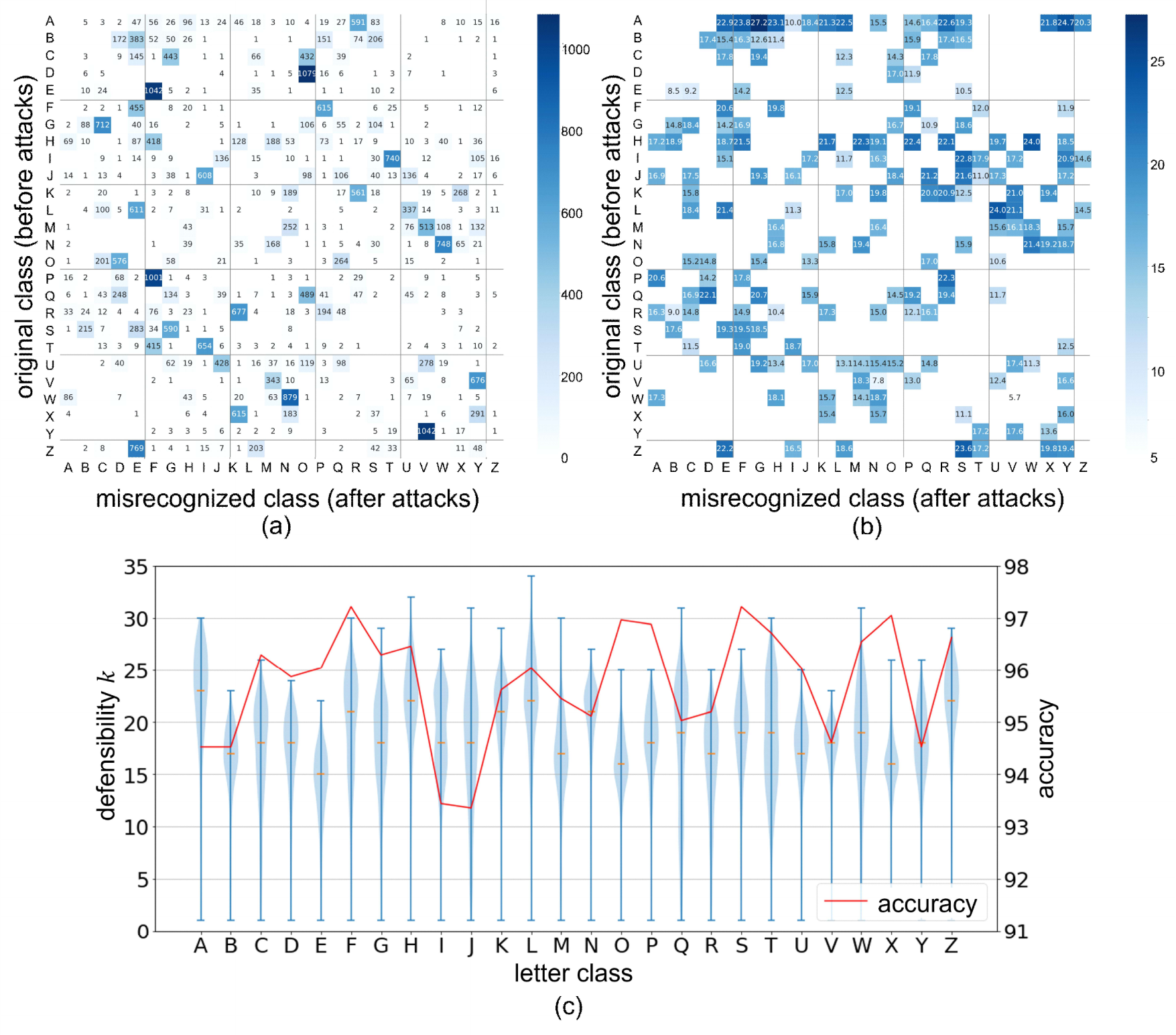}\\[-4mm]
\caption{Pair-wise summaries of the attacking experiment and the distribution of defensibility $k$ of each letter class. (a) shows class confusion matrix, (b) shows average defensibility, and (c) shows the distribution of defensibility $k$ and test classification accuracy. The average defensibility (b) is shown only for the class pairs with more than 10 misrecognized images in (a). The red line graph shows the test classification accuracy of each class before attacks in (c).} \label{fig:matrices}\vspace{-4mm}
\end{figure}

\subsubsection{Class-Wise Analysis of Defensibility}
Fig.~\ref{fig:matrices}(c) shows a violin plot showing the distribution of the defensibility $k$ for each letter class. The median, minimum, and maximum $k$ values are plotted for each class as short horizontal lines. The test classification accuracy before attacks is also plotted as a red line.  
\par
The median defensibility values are around $15\sim 25$ for most classes.
Among 26 classes, \letter{E} shows the lowest median defensibility, although its classification accuracy is not low --- like this, this plot indicates no clear class-wise correlation between the classification accuracy and defensibility. 
It is interesting to observe that the range of frequent $k$ values is narrow for \letterc{N} \letterc{V} and \letter{X} and wide for \letterc{J} \letterc{Q} \letterc{T} and \letterp{W} Namely, the defensibility of the former classes is less variable by font styles, and that of the latter classes is more.\par
\subsubsection{Class-Pair-Wise Analysis of Defensibility}
Fig.~\ref{fig:matrices} shows two matrices for class-pair-wise analysis of defensibility. Fig.~\ref{fig:matrices}(a) shows a class confusion matrix. As expected, character images are misrecognized as their neighboring class by attacks. For example, the similar class pairs, such as  \letter{D}$\to$\letterc{O} \letter{E}$\to$\letterc{F}
\letter{P}$\to$\letterc{F} and 
\letter{Y}$\to$\letterc{V} have more misrecognition pairs.
This fact simply confirms that misrecognitions after repetitive attacks occur not randomly but in a similar class.  
\par
%
Although the confusion matrix is roughly symmetric, careful observation reveals many asymmetric cases. For example, \letter{B}$\to$\letter{E} is frequent (383 images), but \letter{E}$\to$\letter{B} is rare (10 images). Other examples are 
\letter{A}$\leftrightarrow$\letter{R} (591 and 33), \letter{H}$\leftrightarrow$\letter{F} (418 and 20), 
\letter{L}$\leftrightarrow$\letter{U} (337 and 16), and
\letter{T}$\leftrightarrow$\letter{F} (415 and 25). These asymmetric cases inherit the asymmetric property of the nearest-neighbor relationship; the nearest neighbor class of \letter{A} will be \letterc{R} and that of \letter{R} will not be \letter{A} but \letterp{K}
\par

Fig.~\ref{fig:matrices}(b) shows the average defensibility of each class pair. For example, the average defensibility of the class pair \letter{A}$\to$\letter{E} is $22.9$, which is the average number of attack operations on the 47 samples of \letter{A} misrecognized as \letter{E}. Simply speaking, 
an average of $22.9$ attacks is required to misrecognize \letter{A} as \letter{E}.
Note that the average defensibility is not shown in (b) when the number of misrecognized images is less than 10 for reliable analysis.\par
Like Fig.~\ref{fig:matrices}(a), this matrix (b) has a nonzero value for a similar class pair. 
In addition, it is roughly symmetric but still includes asymmetric pairs (such as \letter{E}$\leftrightarrow$\letter{F} (14.2 and 20.6)), like Fig.~\ref{fig:matrices}(a). The class pairs with large shape differences, such as
\letter{A} $\leftrightarrow$\{\letterc{G} \letter{F}\} and \letter{B} $\leftrightarrow$ \letterc{F} tend to have large average defensibility.\par
Interpretation of the average defensibility values is neither straightforward nor intuitive. For example, a similar class pair \letter{O}$\to$\letter{Q} needs 17.0 attacks, whereas
\letter{B}$\to$\letter{H} only needs 11.4 attacks. This difficulty in interpretation is thought to be because defensibility varies greatly depending on the position within the intra-class distribution.
\par
%
As noted above, the average defensibility matrix of Fig.~\ref{fig:matrices}(b) is similar to the confusion matrix (a); however, they represent different relationships between each class pair. If the element of $c\to c'$ is large (say, 800) in the confusion matrix (a), the class $c$ includes many (i.e., 800) images whose nearest neighbor class is $c'$. On the other hand, if the element of $c\to c'$ is large in the average defensibility matrix (b), the 800 images in $c$ are far from the class boundary between $c$ and $c'$ on average. In short, the matrix (a) shows a sample count, whereas (b) a distance. 
These different meanings of the matrices cause inconsistent relationships between class pairs. For example, \letter{E}$\to$\letter{F}(1042) $>$\letter{F}$\to$\letter{E}(455) in (a), 
whereas \letter{E}$\to$\letter{F}(14.2) $<$ 
\letter{F}$\to$\letter{E}(20.6) in (b).


\section{Estimating Defensibility by Regression\label{sec:regression}}


\subsection{Deep Regression Model for Defensibility Estimation}
To answer Question 2 in Section~\ref{sec:intro} ({\it ``Can we estimate how defensive a given letter image is before attacks?''}), we try to estimate the defensibility of each original (i.e., non-attacked) letter image by a deep regression model (without actual attacks). If we can develop a regression model with high accuracy, it will indicate a correlation between the letter shape and its defensibility. 
\par
A deep regression model is trained for each of the 26 letter classes, and its details are as follows. The model is a simple CNN with the same structure as for letter classification presented in the previous section, except for the number of output units ($26\to 1$). Then, the CNN model is trained to estimate the defensibility of the input letter image with the standard MSE loss and AdaDelta optimizer. As noted above, the input image is an original (i.e., non-attacked) image. The ground-truth of the defensibility $k$ is determined by the experimental result of the previous section.  The same criterion as the previous experiment is used to terminate the training process.
\par

In this experiment, we use about 28,700 images of the test set for Section~\ref{sec:attack}. This is because we know their defensibility $k$ through the attack experiment of the previous section. We split them into training, validation, and test sets for the experiment in this section. Consequently, for each model (i.e., for each letter class), these sets contain about 800, 100, and 200 images, respectively. \vspace{-2.5mm}
\begin{figure}[t]
\centering
\includegraphics[width=\textwidth]{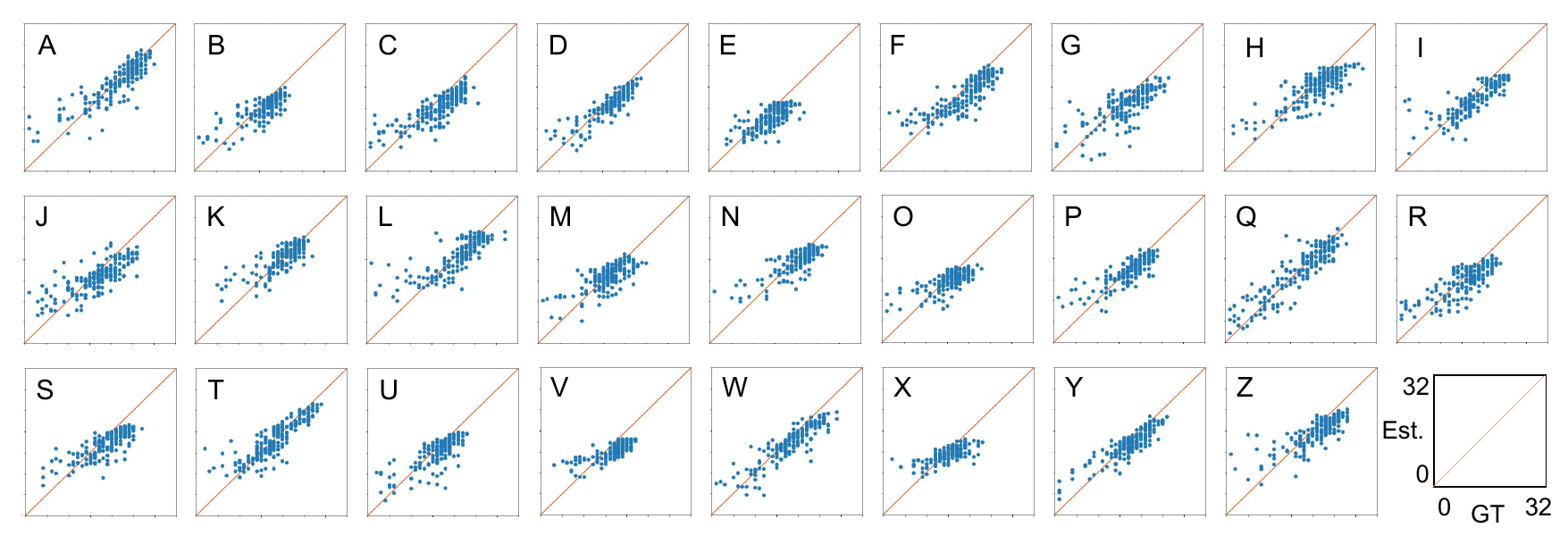}\vspace{-4mm}
\caption{Defensibility estimation result for each class.} \label{fig:estimation-y-y-plot}
\vspace{-4mm}\end{figure}

\subsection{Estimation Experiment}
Fig.~\ref{fig:estimation-y-y-plot} shows the estimation result for each letter class as a $y$-$y$ plot. Its horizontal axis corresponds to the ground-truth, i.e., the defensibility $k$ measured by the attack experiment in Section~\ref{sec:attack}, and the vertical axis to the estimated defensibility. From these plots, we found a clear positive correlation between ground-truth and estimation. This fact proves that there are strong correlations between letter shapes and defensibility. 
\par
Another observation is that the defensibility of fragile letters is often overestimated\footnote[1]{Assume a regression problem regarding $y=f(x)$. When $y$ is non-negative, it is well-known that the resulting $f$ tends to show such overestimation ($y>0$) for the sample $x$ whose $y\sim 0$. (By the non-negativity, $y$ fluctuates only in positive directions and never becomes negative.) However, the $y$-$y$ plots in Fig.~\ref{fig:estimation-y-y-plot} show more significant overestimation around $y\sim 0$.}. As noted above, fragile letters are often printed in some decorative fonts and thus show large variations. On the other hand, 
the number of fragile letters is not large. Consequently, the regression model failed to learn such fragile letters accurately and gave erroneous defensibility estimates.
\par

\section{Generating Defensive Letter Images}

\subsection{Defensive Letter Image Generation by GAN}

\begin{figure}[t]
\centering
\includegraphics[width=0.9\textwidth]{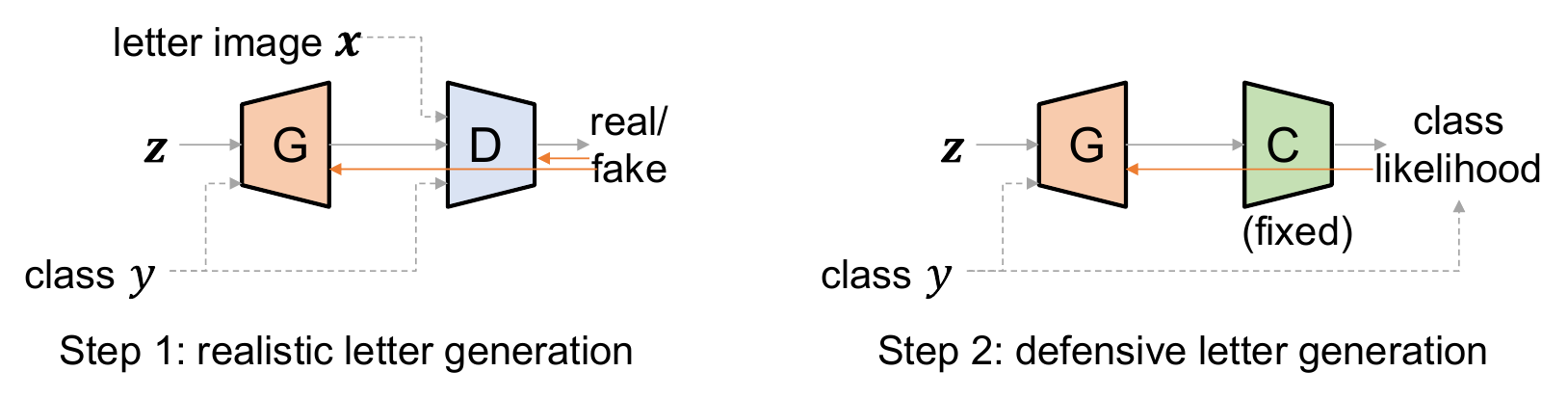}\\[-5mm]
\caption{Defensive letter image generation by GAN.} \label{fig:GAN-model}\vspace{-4mm}
\end{figure}

To answer Question 3 in Section~\ref{sec:intro} ({\it ``Can we control the letter images to be more defensive against adversarial attacks?''}), we try to generate letter images with higher defensibility. For this purpose, we propose a cGAN-based image generation model. Its technical highlight is a two-step training process, as shown in Fig.~\ref{fig:GAN-model}. The first step is a standard GAN training process for generating realistic letter images, where we train a simple cGAN with a pair of a generator and a discriminator, and the discriminator tries to detect fake images by the generator. We adopt the binary cross entropy loss for this step. Specifically, loss functions for the generator $G$ and the discriminator $D$ are as follows:\vspace{-1mm}
\begin{align}
    {\cal L}_G &= \mathbb{E}_{(\mathbf{z}, y)\sim p(\mathbf{z},y)}[-\log D(G(\mathbf{z},y),y)],\\
    {\cal L}_D &= \mathbb{E}_{(\mathbf{x}, y)\sim p(\mathbf{x},y)}[\log D(\mathbf{x},y)] + \mathbb{E}_{(\mathbf{z},y)\sim p(\mathbf{z},y)}[1-\log D(G(\mathbf{z},y),y)],
\end{align}
where $\mathbf{x}$ is an input image, $y$ is a class label, $\mathbf{z}$ is a random noise vector, $G(\mathbf{z},y)$ is a generator that outputs an image of class $y$ from noise vector $\mathbf{z}$, and $D(\mathbf{x},y)$ is the probability that the input image
$\mathbf{x}$ of class $y$ is real (i.e., non-generated) rather than fake (i.e., generated).
\par
Then, in the second and more important step, the discriminator is replaced with a letter image classifier $C$ so that the generator can generate more defensive letter images. Note that we use the CNN $C$ prepared as a classifier to be attacked in Section~\ref{sec:attack-experiment}, and all the parameters of $C$ are frozen during this step. The loss function for this step is the following negative log-likelihood:\vspace{-2mm}
\begin{equation}
    {\cal L}_C  = \mathbb{E}_{(\mathbf{z},y)\sim p(\mathbf{z},y)}[-\log C_y(G(\mathbf{z},y))],
    \label{eq:defensive}\vspace{-2mm}
\end{equation}
where $C_y(\mathbf{x})$ is the $y$-th element of the logit vector obtained from the classifier. 
The idea behind this loss function Eq.(\ref{eq:defensive}) is that we want to train the generator to output the character images that can minimize the classification loss; in other words, the generated letter images will become more easily recognized by $C$. It is an operation opposite to the attacking operation by Eq.~(\ref{eq:FGSM}), and therefore the generated images will have higher defensibility.
\par

\begin{figure}[t]
\centering
\includegraphics[width=\textwidth]{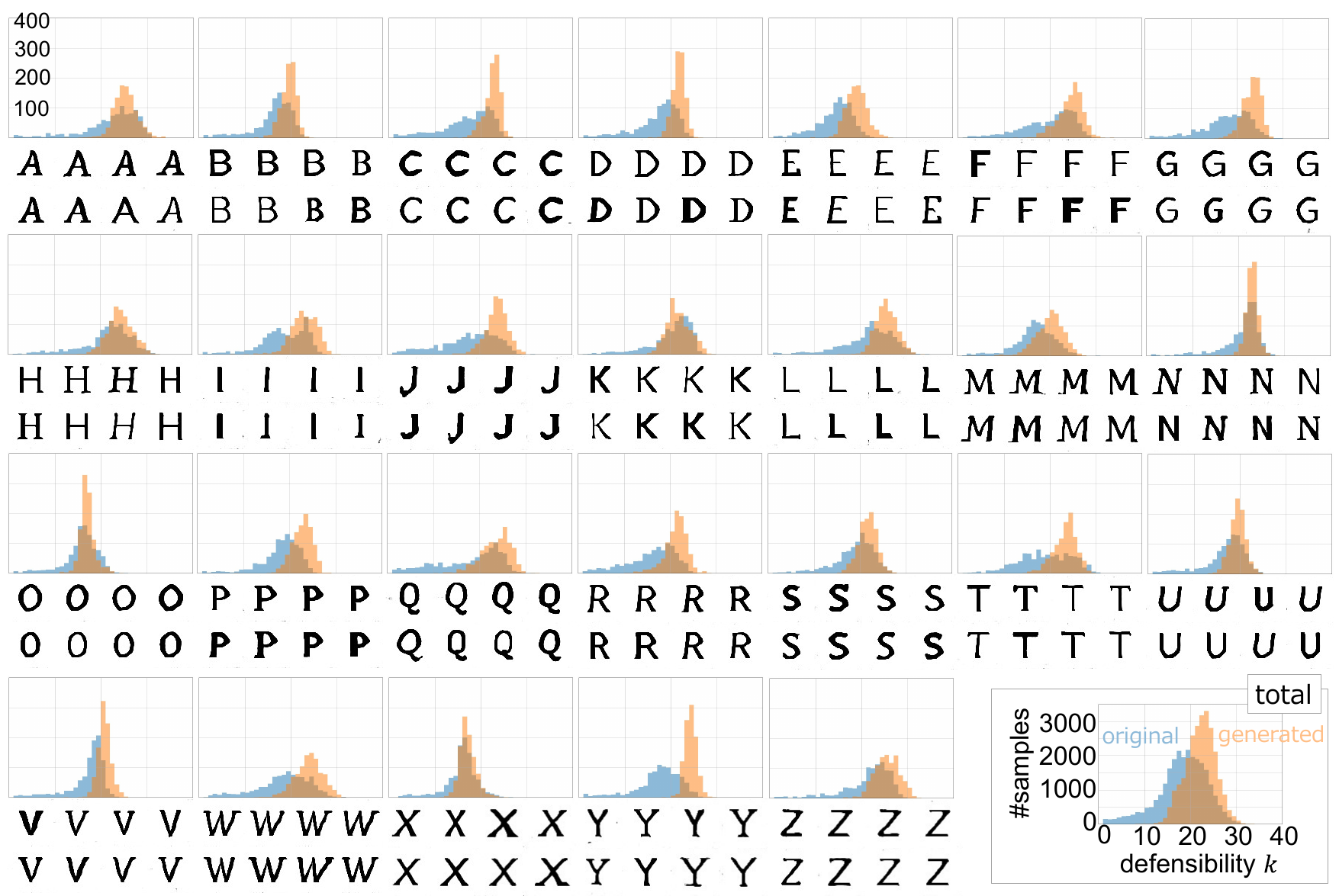}
\caption{Generation results of defensive letter images. For each letter, the histogram of defensibility for 1,000 original images (blue) and 1,000 generated images (orange) are shown along with the generated images with the top eight defensibilities.} \label{fig:generated-examples}
\clearpage
\vspace{-6mm}
\end{figure}
\subsection{Generation Experiment}
\subsubsection{Quantitative Evaluation of Generated Results}
Fig.~\ref{fig:generated-examples} shows the defensibility histogram of each letter. 
Each histogram shows the defensibility distributions of 1,000 original and 1,000 generated images. In addition to those 26 histograms, the total histogram of 26 letters is shown in the lower right corner. The defensibility $k(\mathbf{x})$ of each image $\mathbf{x}$ is counted by actually attacking the original (or generated) image $\mathbf{x}$ by I-FGSM, like the experiment in Section~\ref{sec:attack}. In other words, the defensibility here is {\em not} an estimation $\hat{k}(\mathbf{x})$ by the regression model of Section~\ref{sec:regression}.\par

First, we can see a successful result that the generated images by the GAN-based model of Fig.~\ref{fig:GAN-model} have more defensible than the original images. The average defensibilities of the original and generated images were 17.8 and 21.7, respectively. This means four more attacks are necessary to misrecognize the generated images on average. Such an increase in defensibility in the generated images can be seen in all classes. In addition, none of the generated images have defensibility of less than 10.\par

Second, the maximum defensibility of the generated images is not significantly higher than that of the original images (except for \letterc{E} \letterc{I} \letterc{J} and \letter{W}). This suggests that the current GAN model of Fig.~\ref{fig:GAN-model} has difficulty generating character images that have never been seen, even though it tries to increase the defensibility.
\subsubsection{Qualitative Evaluation of Generated Results}
Fig.~\ref{fig:generated-examples} also shows the generated images with the top eight defensibilities for each class. Therefore, these images correspond to the rightmost side of the orange histogram of the generated image. 
As noted above, these generated images have defensibility equal to or greater than the maximum defensibility of the original images. \par
Despite their higher defensibility, their style is neutral (except for tiny fluctuations in the stroke contour) and not very decorative. This result coincides with Fig.~\ref{fig:attacked-examples}, where less decorative images have higher defensibility. One more observation is that their style is similar to each other; they keep higher defensibility at the cost of variety. In summary, both the discriminative approach (to the first research question) and the generative approach (to the third question) confirmed that standard characters are defensive.

\section{Discussion\label{sec:discussion}}
\subsection{What Are Reasonable Attacks?\label{sec:discussion-attack}}
In this paper, we do not assume any constraint on the attacking process by I-FGSM and the attacked images. However, we may introduce  some constraints to regulate the attacked images. In fact, a popular attacking method, called Projected Gradient Descent (PGD)~\cite{Makelov2018}, projects the attacked image to a prespecified image subspace. For example, we can use the subspace of all binary images so that the adversarial letter images become binary. Although it seems a reasonable constraint, our preliminary trial proves that the resulting images often showed extra black pixels or black connected components in the background region, independently of letter shapes. To suppress them, we needed to introduce another extra constraint besides the binary constraint.\par
However, it is also true that there are no {\em reasonable} constraints that everyone can agree on. In other words, tuning the constraint $Q$ would give adversarial images strongly biased by our prejudices towards our own ``ideal'' attacked letters. Since this paper is the first attempt to know the defensibility of letter images, we try not to optimize such a constraint to be free from our prejudices. In future work, we will design constraints $Q$ that fit a specific application scenario.\par

\begin{figure}[t]
\centering
\includegraphics[width=0.8\textwidth]{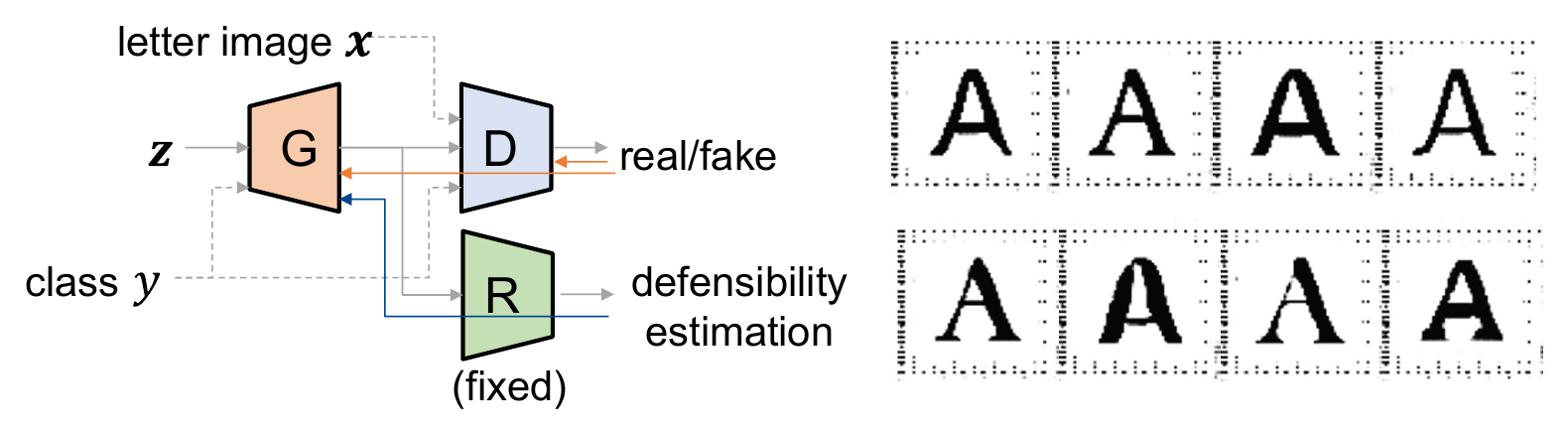}\\[-3mm]
\caption{A GAN-based defensive letter generation model that did {\em not} work as expected.  Eight \letter{A} images are the generated images by this model and show noisy dots around their image boundary.} \label{fig:multi-task-GAN}\vspace{-5.5mm}
\end{figure}

\subsection{Why Is the Regression Model {\em not} Used with GAN?\label{sec:discussion-GAN}}
One might think that the regression model is useful in the GAN-based defensive letter generation. For example, we can consider another GAN-based defensive letter generation model like Fig.~\ref{fig:multi-task-GAN} instead of Fig.~\ref{fig:GAN-model}. In this model, the regression model prepared in Section~\ref{sec:regression} is added to estimate the defensibility of a generated letter. Then, the model provides a  gradient to the generator to generate letters with higher defensibility.
\par
However, in the current setup, this model does not work as expected. As explained in Section~\ref{sec:regression}, the regression model is trained with the letter images attacked I-FGSM. In other words, the model does not know
the images other than these attacked images. Therefore, for example, if the generator generates strange noisy images like \letter{A} shown in the right part of Fig.~\ref{fig:multi-task-GAN}, the model might give overestimated defensibility for them. In fact, we measured the true defensibility of these \letter{A}s by I-FGSM and found that their true defensibility is much lower than the estimated one. As this result, unfortunately, the generator continues to generate letter images like these \letter{A}s, which actually have low defensibility. A naive remedy is to update the regression model jointly with the newly generated images and their real defensibility;  our preliminary trial, however, found this training strategy is unstable. Consequently, we need to develop a totally new framework other than Fig.~\ref{fig:multi-task-GAN} as future work.
\section{Conclusion and Future Work}\vspace{-0.4mm}

Unlike the standard adversarial defense research, where classifiers are the target to be defended, our purpose is to defend classification targets, i.e., letter images. For this purpose, we conducted various experiments and analyses for the three tasks: (1)~Measurement of the actual defensibility of letter images. 
(2)~Estimation of the defensibility from letter images. 
(3)~Generation of letter images with higher defensibility. From (1), we confirmed that letters in simpler and more standard fonts tend to have higher defensibility. From (2), we confirmed a close relationship between the letter shape and the defensibility. Finally, from (3), we confirmed that a GAN-based model with a classifier could generate letter images with higher defensibility.\par
To the best of the authors' knowledge, this is the first attempt to observe the defensibility of letter images, and thus, as discussed in Section~\ref{sec:discussion}, there are many open questions regarding further analyses of the letter defensibility. The design of adversarial attack algorithms and constraints during the attack is an important future direction because it determines not only the defensibility of each letter image but also the appearance of the generated defensive letter images. In addition to the current bitmap-based generation, we may generate letter contours (in, for example, TrueType format). Also, we are going to realize a generation model that transforms a given letter image to be more defensive.
\par
Throughout this paper, we have seen the defensibility of letter images against adversarial attacks to non-human classifiers (i.e., CNN-based classifiers). On the other hand, as noted in Section~\ref{sec:intro}, letter images have high defensibility against various natural attacks, such as perspective distortions, handwriting fluctuations, partial occlusion, and decorations, to human classifiers. If we can tie together these various findings about the defensibility of letter images for computers and humans, we may find clues to understanding why and how letter images (i.e., alphabets) are designed to retain their readability against various distortions (i.e., attacks).
\section*{Acknowledgment}
This work was supported in part by JSPS KAKENHI Grant Numbers JP22H00540.

%
%
%
\bibliographystyle{splncs04}
\bibliography{references.bib}

\begin{thebibliography}{10}
\providecommand{\url}[1]{\texttt{#1}}
\providecommand{\urlprefix}{URL }
\providecommand{\doi}[1]{https://doi.org/#1}

\bibitem{Akhtar2018}
Akhtar, N., Mian, A.: Threat of adversarial attacks on deep learning in
  computer vision: A survey. IEEE Access  \textbf{6},  14410--14430 (2018)

\bibitem{Chakraborty2018}
Chakraborty, A., Alam, M., Dey, V., Chattopadhyay, A., Mukhopadhyay, D.:
  Adversarial attacks and defences: A survey. arXiv preprint arXiv:1810.00069
  (2018)

\bibitem{Goodfellow2015}
Goodfellow, I.J., Shlens, J., Szegedy, C.: Explaining and harnessing
  adversarial examples. In: Proceedings of the 3rd International Conference on
  Learning Representations (ICLR) (2015)

\bibitem{Kurakin2017}
Kurakin, A., Goodfellow, I.J., Bengio, S.: Adversarial examples in the physical
  world. In: Proceedings of the 5th International Conference on Learning
  Representations (ICLR) (2017)

\bibitem{Long2022}
Long, T., Gao, Q., Xu, L., Zhou, Z.: A survey on adversarial attacks in
  computer vision: Taxonomy, visualization and future directions. Computers and
  Security  \textbf{121} (2022)

\bibitem{Machado2021}
Machado, G.R., Silva, E., Goldschmidt, R.R.: Adversarial machine learning in
  image classification: A survey toward the defender's perspective. ACM
  Computing Surveys  \textbf{55}(1),  1--38 (2021)

\bibitem{Makelov2018}
Makelov, A., Schmidt, L., Tsipras, D., Vladu, A., Science, C.: Towards deep
  learning models resistant to adversarial attacks. In: Proceedings of the 6th
  International Conference on Learning Representations (ICLR) (2018)

\bibitem{Kaviani2022}
Sara, K., Ki, J.H., Insoo, S.: Adversarial attacks and defenses on {AI} in
  medical imaging informatics: A survey. Expert Systems with Applications
  \textbf{198},  116815 (2022)

\bibitem{uchida2007}
Uchida, S., Sakai, M., Iwamura, M., Omachi, S., Kise, K.: Extraction of
  embedded class information from universal character pattern. In: Proceedings
  of the 9th International Conference on Document Analysis and Recognition
  (ICDAR). pp. 437--441 (2007)

\bibitem{Wu2021}
Weibin, W., Yuxin, S., Michael, R.L., Irwin, K.: Improving the transferability
  of adversarial samples with adversarial transformations. In: Proceedings of
  the IEEE/CVF Conference on Computer Vision and Pattern Recognition (CVPR).
  pp. 9024--9033 (2021)

\bibitem{Wang2021}
Xiaosen, W., Kun, H.: Enhancing the transferability of adversarial attacks
  through variance tuning. In: Proceedings of the IEEE/CVF Conference on
  Computer Vision and Pattern Recognition (CVPR). pp. 1924--1933 (2021)

\bibitem{Xu2020}
Xu, H., Ma, Y., Liu, H.C., Deb, D., Liu, H., Tang, J.L., Jain, A.K.:
  Adversarial attacks and defenses in images, graphs and text: A review.
  International Journal of Automation and Computing (IJAC)  \textbf{17}(2),
  151--178 (2020)

\end{thebibliography}

\end{document}